\documentclass{article}

    \usepackage[preprint]{tackling_climate_workshop_style}

\usepackage[utf8]{inputenc} 
\usepackage[T1]{fontenc}    
\usepackage{hyperref}       
\usepackage{url}            
\usepackage{booktabs}       
\usepackage{amsfonts}       
\usepackage{nicefrac}       
\usepackage{microtype}      
\usepackage{graphicx}       
\usepackage{float}          
\usepackage{subcaption}     
\usepackage{svg}            
\usepackage{booktabs}       
\usepackage{multirow}

\title{Health Care Waste Classification Using Deep Learning Aligned with Nepal\textquotesingle s Bin Color Guidelines}

\author{%
  Suman Kunwar \\
  DWaste, USA\\
  \texttt{sumn2u@gmail.com} \\
  \And
  Prabesh Rai \\
  Lambton College, Canada \\
  \texttt{raiprabesh775@gmail.com} \\
}

\begin{document}

\maketitle

\begin{abstract}
The increasing number of Health Care facilities in Nepal has added up the challenges on managing health care waste (HCW). Improper segregation and disposal of HCW leads to contamination, spreading  of infectious diseases and risk for waste handlers. This study benchmarks the state of the art waste classification models: ResNeXt-50, EfficientNet-B0, MobileNetV3-S, YOLOv8-n and YOLOv5-s using stratified 5-fold cross-validation technique on combined HCW data. YOLOv5-s achieved the highest accuracy (95.06\%) but fell short with the YOLOv8-n model in inference speed with few milliseconds. The EfficientNet-B0 showed promising results of 93.22\% accuracy but took the highest inference time. Following a repetitive ANOVA test to confirm the statistical significance, the best performing model (YOLOv5-s) was deployed to the web with bin color mapped using Nepal's HCW management standards. Further work is suggested  to  address data limitation and ensure localized context.
\end{abstract}

\keywords{health care waste, deep learning, waste classification, computer vision, Nepal health care waste standards}

\section{Introduction}

Managing HCW is a pressing challenge fueled by increasing waste generation by health care facilities. Improper segregation and disposal of HCW results in the spread of infectious diseases, contamination and puts risk towards waste handlers. Traditionally, HCW are done manually which is often labor-intensive, prone to human errors and poses a threat to  waste handlers. Numerous researchers have demonstrated the feasibility of AI powered solutions for automated waste sorting and its management. 

Studies have applied deep learning (DL)  ResNeXt-50 \cite{zhou_deep_2022} and EfficientNet\_B7 \cite{kumar_artificial_2021} models, with transfer learning to get higher accuracy results in HCW classifications, even with small datasets. The intersection of IoT with models like YOLOv5 \cite{lahoti_multi-class_2024}, YOLOv8 \cite{moktar_medical_2025} and MobileNetV3 \cite{sarkar_modernized_2023} has shown promising results on automated sorting of medical waste showcasing real world implementation.

In the study, Zhou et al. used ResNeXt-50 and achieved 97.2\% accuracy on an eight-class private dataset, while other fine tune transfer learning on EfficientNet-B7,  reaching an accuracy of 99\% on six class dataset. Automated sorting with the help of sensors, robotic arms and edge devices has also shown promising results. These approaches have reported with 97.93\% accuracy highlighting the real-world deployment in healthcare environments \cite{karmakar_ai_2024}. Moktar et al. used YOLO combined with IoT for the detection of HCW and achieved a mAP of 98\% using 6,7860 images. 

With the promising results, challenges like small dataset size, poor image quality and controlled environment testing persists raising concerns \cite{bian_medical_2021} on the scalability, infrastructure feasibility and adoption with existing waste management systems. These put forward the needs of expansion of dataset to generalize the detection, needs of real time deployment with real world settings and integration of AI with IoT for automation.

\begin{figure}[H]
  \centering
  \includegraphics[width=0.97\textwidth]{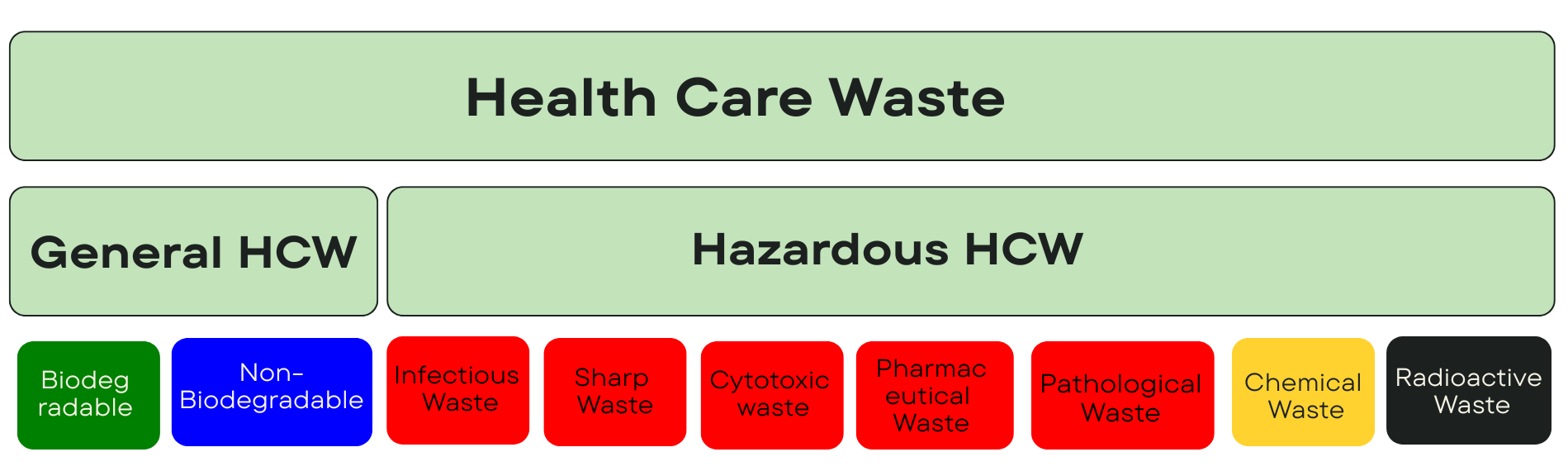}
  \caption{Color codes for waste segregation}
  \label{fig:HCW-Standards}
\end{figure}

In Nepal, there are 16,611 health care facilities \cite{mohp2020_nhcw_sop_misc} serving people that follows to the guidelines set by National Health Care Waste Management Standards and Operating Procedures to manage HCW. These guidelines divide waste in General Health Care Waste (General HCW) and Hazardous Health Care Waste (Hazardous HCW). The Hazardous HCW   spans in multiple categories and is mapped with various color codes as depicted in Figure \ref{fig:HCW-Standards}.

The mapping of individual items based on waste category, label and color codes are shown in Table \ref{tab:waste_categories}.
\renewcommand{\arraystretch}{1.2} 
\begin{table}[H]
\centering
\caption{Healthcare Waste Categories, Types, Color Codes, and Items}
\label{tab:waste_categories}
\resizebox{\textwidth}{!}{%
\begin{tabular}{@{}p{3cm} p{3.5cm} p{2.5cm} p{6.5cm}@{}}
\toprule
\textbf{Waste Category} & \textbf{Type} & \textbf{Color Code} & \textbf{Items} \\ 
\midrule
\multirow{2}{*}{Non-risk HCW} 
  & Biodegradable & Green & Organic wastes; peels of vegetables and fruits; rotten or stale foods \\ 
  & Non-Biodegradable & Blue & Paper (A4, duplex, medicine covers, normal papers); plastic (polythene bags, wrappers, packaging plastic materials, cling wraps); bottles (water, soft drinks); vials; saline bottles \\ 
\midrule
\multirow{7}{*}{Risk HCW} 
  & Infectious & Red & Cotton, gauze, used gloves, IV sets, blood bags, urine bags, used PPE, diapers, pads, sputum containers, collection tubes, dialyzer \\ 
  & Pathological &  & Body organs or tissues; placenta \\ 
  & Cytotoxic &  & Unused or expired cytotoxic drugs; chemotherapy agents \\ 
  & Pharmaceutical &  & Expired medicines \\ 
  & Sharps &  & Needles, blades, scalpels, broken glass, lancets, suture needles \\ 
  & Chemical & Yellow & Discarded chemicals, reagents, solvents, disinfectants \\ 
  & Radioactive & Black & X-ray films; radioisotopes \\ 
\bottomrule
\end{tabular}
}
\end{table}

\begin{figure}[H]
  \centering
  \includegraphics[width=0.90\textwidth, height=5cm]{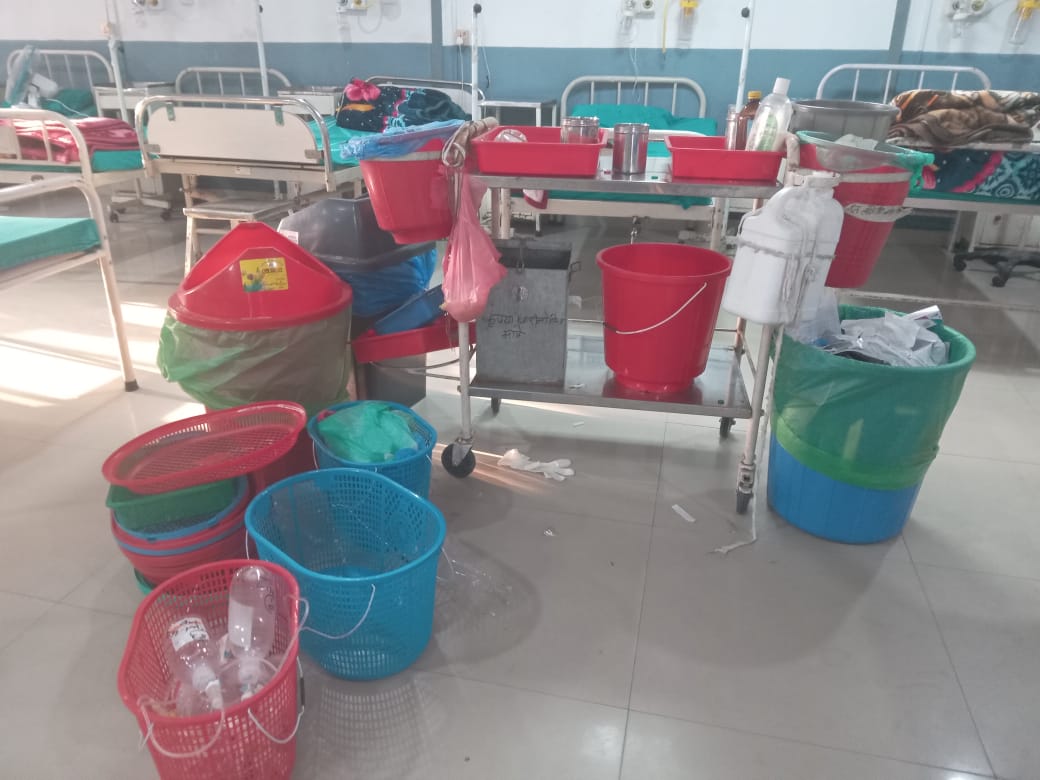}
  \caption{Waste bins at Bhaktapur Cancer Hospital, Nepal}
  \label{fig:Waste_bins}
\end{figure}

A study by Patel et al. on HCWM practices and knowledge in 10 government district hospitals in Madhesh Pradhesh found that the waste handlers showed low adherence to HCWM guidelines and lacked sufficient knowledge, despite having adequate facilities for managing HCM \cite{patel_knowledge_2024}. This poses a risk towards waste handlers and residents nearby hospitals as highlighted by Karki et al. study \cite{karki_risk_2020}. Furthermore, some studies suggest that with proper sorting the amount of HCW can be reduced in Nepal's context \cite{pathak_capacity_2021}. An example of waste bins from one of the hospitals is shown in Figure \ref{fig:Waste_bins}.

The studies demonstrate the effectiveness of DL models on HCW segregation, suggesting that proper sorting can help in waste minimization and de-risk waste handlers. This study, therefore, explores state of the art waste classification models: YOLOv8-n, MobileNetV3-S,  ResNeXt-50, EfficientNet-B0 and YOLOv5-s using stratified K-fold cross-validation method, which is preferred for optimizing sample selection and robust prediction results \cite{pachouly_systematic_2022}. The best performing model mapped with Nepal bin color guidelines is deployed for public usage.

\section{Materials and Methods}
The section discusses the dataset used in this study and the methods used to benchmark the models.

\subsection{Dataset and Preprocessing}
We use a combined dataset: Medical Waste Dataset 4.0 \cite{bruno_medical_2023} and Pharmaceutical and Biomedical Waste dataset \cite{rapeepan_pitakaso_pharmaceutical_2025} to cover major categories of Nepal HCW. Medical Waste Dataset 4.0 was acquired from Tuscany Region, using OAK 4.0 camera device and span across:  gauze, glove pair (latex, nitrile, surgery), glove single (latex, nitrile, surgery), medical cap, medical glasses, shoe cover pair, shoe cover single,  test tube and urine bag. The RGB images are of size 1920 x 1080. The Pharmaceutical and Biomedical Waste dataset is collected by the Engineering UBU and consists of: Body Tissue or Organ, Organic wastes, equipment-packaging (plastic, paper, metal, glass), Syringe needles, Gauze, Gloves, Mask, Syringe and Tweezers.

In the Pharmaceutical and Biomedical Waste dataset, the Gloves contain mixed gloves images, a more generalized class. As the separated types of gloves are already present in the Medical Waste Dataset 4.0, we removed this class to remove the bias. The dataset is divided into 5 stratified folds; each fold containing 80\% train and 20\%  validation data, and can be found in Kaggle\footnote{\url{https://www.kaggle.com/datasets/sumn2u/medical-waste}}. The oversample class images were reduced using the median class count value to handle the synthetic bias. For under sampling classes, data augmentation techniques such as flipping and brightness contrast were used.

\subsection{Model Training}

The experiments were conducted using  two NVIDIA Tesla T4 GPUs. Models were tested using 5-fold stratified cross validation,  with each fold trained for 30 epochs using MobileNetV3-S, ResNeXt-50, EfficientNet-B0 and YOLO models (v5-s/v8-n). The stratified K-fold technique was used to maintain the same percentage of instances for each label \cite{kaliappan_impact_2023}.

\begin{figure}[H]
    \centering
    \begin{subfigure}{\textwidth}
        \centering
        \includegraphics[width=0.85\linewidth]{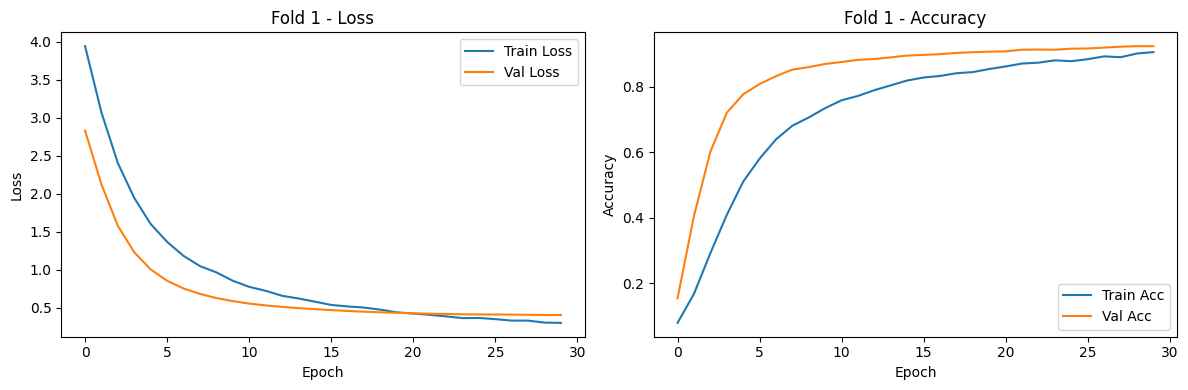}
        \caption{Fold 1}
    \end{subfigure}
    
    \begin{subfigure}{\textwidth}
        \centering
        \includegraphics[width=0.85\linewidth]{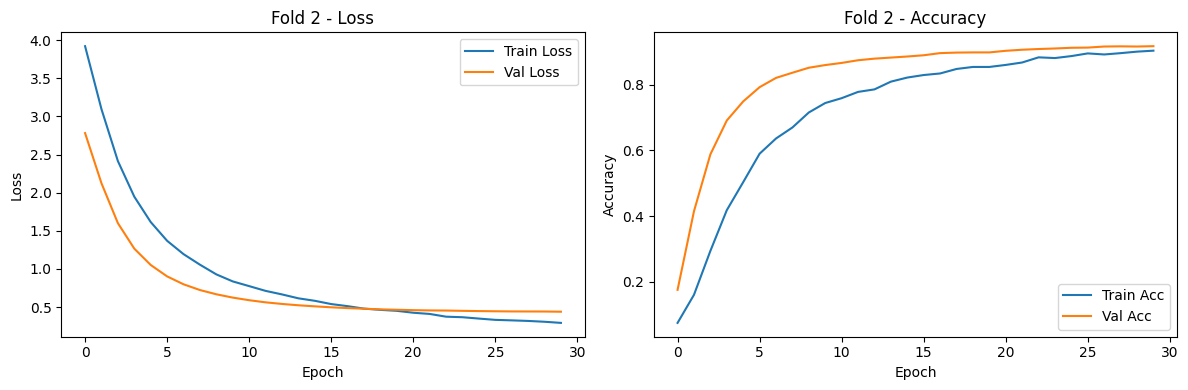}
        \caption{Fold 2}
    \end{subfigure}
    
    \begin{subfigure}{\textwidth}
        \centering
        \includegraphics[width=0.85\linewidth]{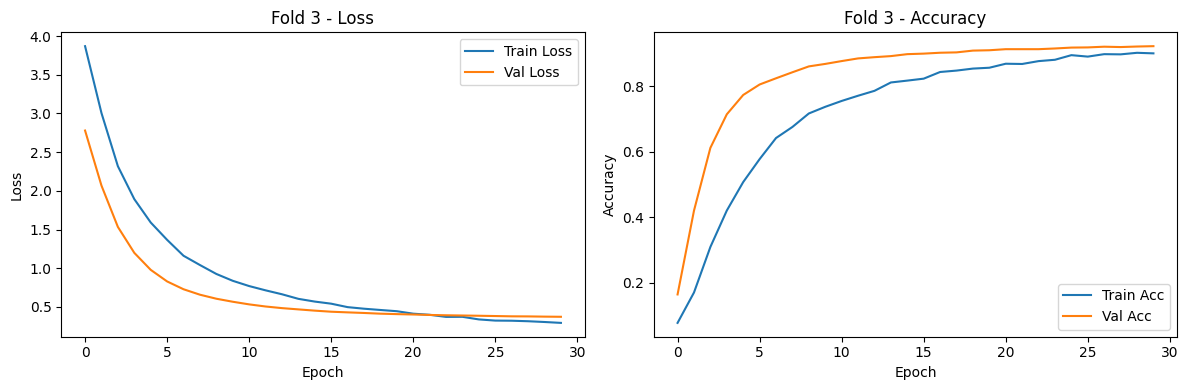}
        \caption{Fold 3}
    \end{subfigure}
    
    \begin{subfigure}{\textwidth}
        \centering
        \includegraphics[width=0.85\linewidth]{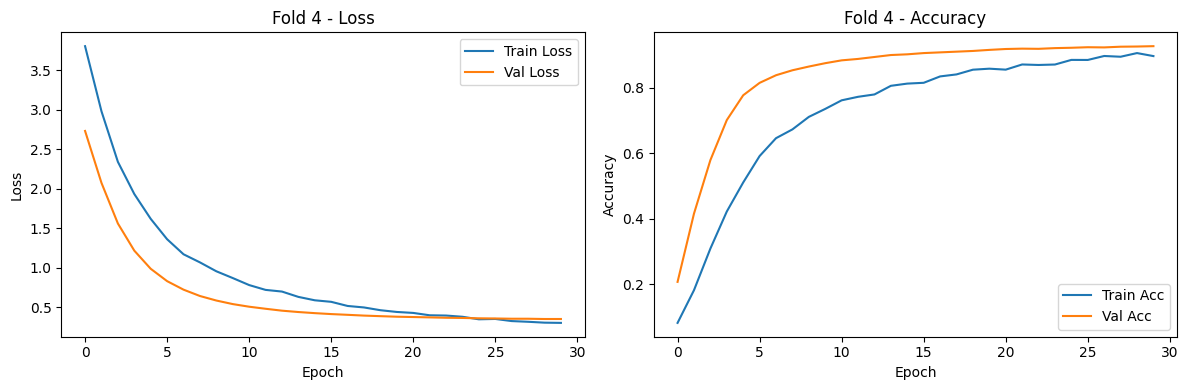}
        \caption{Fold 4}
    \end{subfigure}
    
    \begin{subfigure}{\textwidth}
        \centering
        \includegraphics[width=0.85\linewidth]{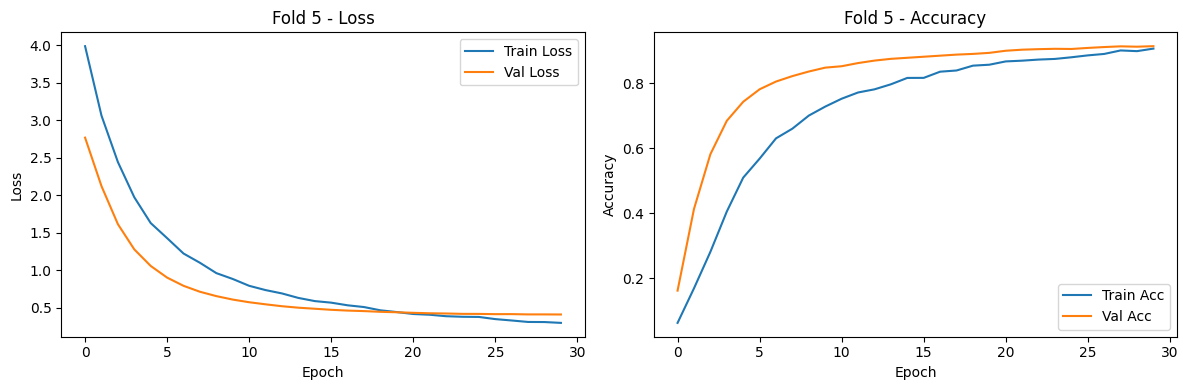}
        \caption{Fold 5}
    \end{subfigure}

    \caption{Training loss and accuracy curves for each fold in 5-fold cross-validation of EfficientNet-B0 model}
    \label{fig:training_curves}
\end{figure}

DL models such as MobileNetV3-S, ResNeXt-50 and EfficientNet-B0, used pre-trained weights trained in ImageNet dataset with frozen base model layers and a custom classification head. In contrast, YOLO models (v5-s/ v8-n) are trained from scratch. The precision, recall, F1-score and accuracy of each model along with inference time are reported as weighted average. Figure \ref{fig:training_curves}. depicts the training and validation loss and accuracy for each fold of EfficientNet-B0 model.

The repeated measures of ANOVA results were also compared to determine statistical significance. When significance was found, Tukey's HSD Post-Hoc test was used to identify differing pairs while controlling the Type I error rate \cite{mchugh_multiple_2011}.

\section{Results and Discussion}

The experiment's results are shown in Table \ref{tab:model_comparison}. YOLOv5-s, YOLOv8-n and EfficientNet-B0 outperformed MobileNetV3-S and ResNeXt-50 in most metrics. YOLOv5-s showed better performance than others but fell behind the YOLOv8-n model in inference time. YOLOv5-s achieved the highest accuracy of 95.06\%; in comparison, the ResNeXt-50 achieved 74.51\% accuracy making it the least accurate. MobileNetV3-s is quite competitive but fell short behind the top performers.

\begin{table}[H]
\centering
\caption{Comparison of performance metrics of various models.}
\label{tab:model_comparison}
\begin{tabular}{lccccc}
\hline
\textbf{Model} & \textbf{Accuracy} & \textbf{Precision} & \textbf{Recall} & \textbf{F1-Score} & \textbf{Inference Time (ms)} \\
\hline
YOLOv8-n        & 0.9468 & 0.9644 & 0.9468 & 0.9457 & \textbf{9.29} \\
MobileNetV3-S   & 0.9105 & 0.9290 & 0.9105 & 0.9095 & 369.24 \\
ResNeXt-50      & 0.7451 & 0.7653 & 0.7451 & 0.7448 & 395.74 \\
EfficientNet-B0   & 0.9322 & 0.9481 & 0.9322 & 0.9304 & 444.67 \\
YOLOv5-s        & \textbf{0.9506} & \textbf{0.9665} & \textbf{0.9506} & \textbf{0.9487} & 10.97 \\
\hline
\end{tabular}
\end{table}

YOLOv8-n and YOLOv5-s showed lower inference time than other models and they achieved better results across accuracy, precision, recall, or F1-score are priorities followed by EfficientNet-B0. In real world applications where inference time is crucial, YOLO (v5-s/ v8-n) models are the best choice. The training time of EfficientNet-B0 is much higher compared to other models along with the model size.

\begin{figure}[H]
  \centering
  \includegraphics[width=0.95\textwidth]{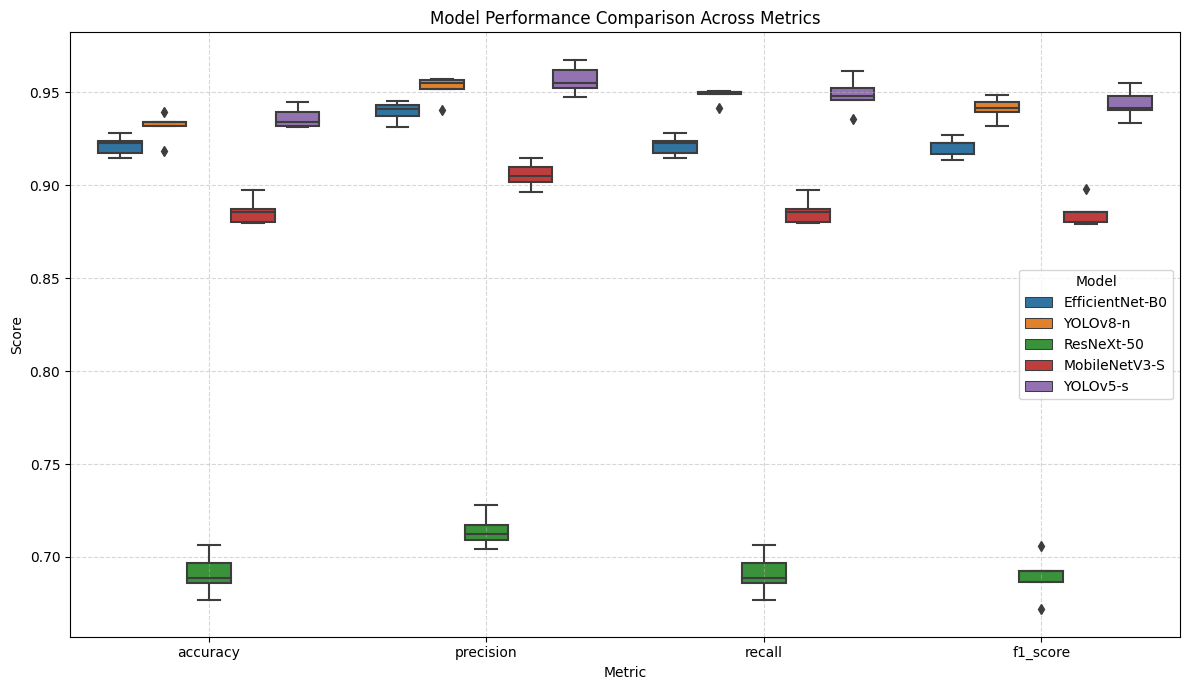}
  \caption{Model Performance Comparison Across Metrics using 5 fold}
  \label{fig:model_comparision_metrics}
\end{figure}

The repeated ANOVA measures across the 5 stratified folds confirmed the highly significant effect of the model on the performance metrics. A significant difference between metrics themselves and the interaction between model and metrics was also highly significant. These results indicate that both the choice of model and type of evaluation metric can influence the performance, suggesting some models perform better on specific metrics than others. The model performance comparison across various models using different metrics can be seen in Figure \ref{fig:model_comparision_metrics}, which clearly shows that the YOLO models were outperforming the rest.

Our results were compared with the study that introduced the Medical Waste Dataset 4.0 shown in Table \ref{tab:study_comparison}. 

\begin{table}[H]
\centering
\caption{Comparison of studies and their best-performing models.}
\label{tab:study_comparison}
\begin{tabular}{lccc}
\hline
\textbf{Study} & \textbf{Number of Classes} & \textbf{Best Model} & \textbf{Accuracy} \\
\hline
Bruno et al. & 7 classes & EfficientNet-B0 & 99.45\% \\
Our Work & 23 classes & YOLOv5-s & 95.06\% \\
\hline
\end{tabular}
\end{table}

While the work by Bruno et al. achieved a high accuracy of 99.45\% with EfficientNet-B0, its scope was limited to only seven classes. Our approach, utilizing the YOLOv5-s model, reached an accuracy of 95.06\% on a significantly more challenging 23-class task, demonstrating robust and scalable performance. These results set a new standard for detailed medical waste sorting and confirm that YOLO models are useful for real-time sorting.

\begin{figure}[H]
  \centering
  \includegraphics[width=0.95\textwidth]{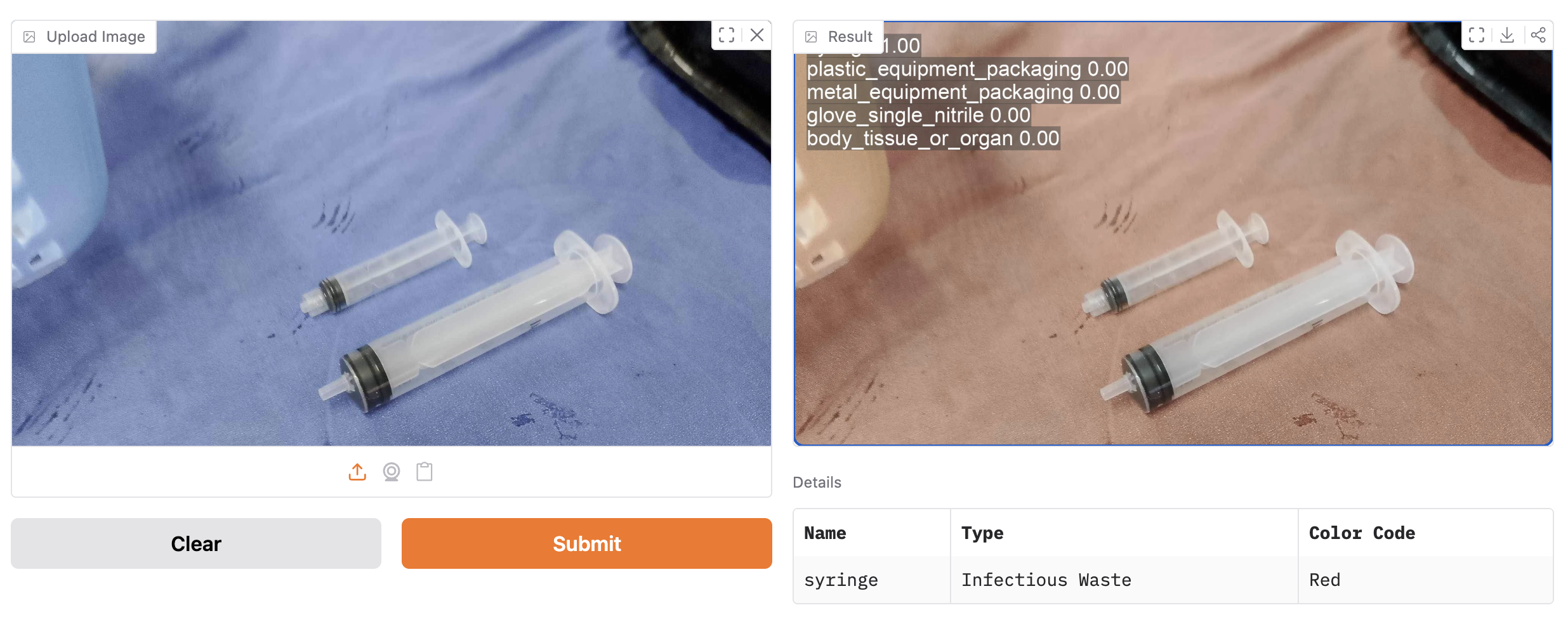}
  \caption{HCW Detection sample app with bin color based on Nepal's guidelines}
  \label{fig:model_deployment}
\end{figure}

From our experiment, we found that YOLOv5-s performed well and is deployed in a Hugging Face space for real-world classification of waste images with bin mapping based on Nepal's guidelines\footnote{\url{https://huggingface.co/spaces/iamsuman/medical-waste-detector}}. Figure \ref{fig:model_deployment} shows a sample HCW image being classified with bin color. 

Our study has some limitations, including missing some classes: cytotoxic, radioactive, soiled and pathological, chemical and liquid waste that are included in Nepal's waste system. Additionally, the combined dataset is skewed towards common items such as gloves and gauze and is collected in non-Nepali settings. There is still a high chance that the model might fail in real world as waste often appears in occluded, mixed, and cluttered  packaging. To mitigate these issues, further work needs to be done especially on data collection that represents generalized data in local settings.

\begin{ack}
We thank Tika Ram Poudel from Waste Service Private Limited for providing HCW images and bins from Bhaktapur Cancer Hospital, Nepal. 
\end{ack}

\bibliographystyle{unsrt}
\bibliography{sample}

\end{document}